\documentclass[letterpaper]{article} 
\usepackage{aaai24}
\usepackage{times}  
\usepackage{helvet}  
\usepackage{courier}  
\usepackage[hyphens]{url}  
\usepackage{graphicx} 
\usepackage{color, colortbl}
\usepackage{subcaption}
\usepackage{amsmath}
\usepackage{amsthm}
\usepackage{amssymb,amsfonts}
\usepackage{dsfont}
\usepackage{arydshln}
\usepackage{natbib}  
\usepackage{caption} 
\usepackage{algorithm}
\usepackage[noend]{algpseudocode}
\usepackage{adjustbox}
\usepackage{tabularx}
\usepackage{multirow}
\usepackage{makecell}
\usepackage{diagbox}
\usepackage{newfloat}
\usepackage{listings}
\usepackage{appendix}
\DeclareCaptionStyle{ruled}{labelfont=normalfont,labelsep=colon,strut=off} 
\floatstyle{ruled}
\newfloat{listing}{tb}{lst}{}
\floatname{listing}{Listing}
\algrenewcommand\algorithmicrequire{\textbf{Input:}}
\algrenewcommand\algorithmicensure{\textbf{Output:}}
\usepackage{xcolor}
\definecolor{Gray}{gray}{0.9}

\newtheorem{manualtheoreminner}{Theorem}
\newenvironment{manualtheorem}[1]{%
  \renewcommand\themanualtheoreminner{#1}%
  \manualtheoreminner
}{\endmanualtheoreminner}

\newtheorem{manuallemma}{Lemma}
\newenvironment{manuallem}[1]{%
  \renewcommand\themanuallemma{#1}%
  \manuallemma
}{\endmanuallemma}

\newtheorem{manualcorollary}{Corollary}
\newenvironment{manualcoroll}[1]{%
  \renewcommand\themanualcorollary{#1}%
  \manualcorollary
}{\endmanualcorollary}

\newtheorem{manualdefinition}{Definition}
\newenvironment{manualdef}[1]{%
  \renewcommand\themanualdefinition{#1}%
  \manualdefinition
}{\endmanualdefinition}

\title{Better Not to Propagate: Understanding Edge Uncertainty and Over-smoothing in Signed Graph Neural Networks}
\author{
    Yoonhyuk Choi$^1$, Jiho Choi$^2$, Taewook Ko$^3$, Chong-Kwon Kim$^4$\\
}
\affiliations{
    Samsung SDS$^1$, KAIST$^2$, Seoul National University$^3$, Korea Institute of Energy Technology$^4$\\

    chldbsgur123@gmail.com, jihochoi@kaist.ac.kr, taewook.ko@snu.ac.kr, ckim@kentech.ac.kr
}

\usepackage{bibentry}

\begin{document}

\maketitle

\begin{abstract} 


Local smoothing, driven by message-passing (MP) under varying homophily levels, is crucial in Graph Neural Networks (GNNs).
Recent research has examined this smoothing effect, which diminishes node separability after MP, by analyzing the expected distribution of node features.
These studies offer theoretical insights into the local smoothing effects of different propagation methods, such as positive, signed, and blocked MPs.
Node separability is theoretically determined by two hyperparameters: a fixed local homophily and a dynamic edge error ratio.
However, previous analyses have constrained these values, potentially limiting optimal MP selection for GNNs. To overcome this, we propose a novel method for estimating homophily and edge error ratios, along with a dynamic selection mechanism between blocked and signed propagation during training.
Our theoretical analysis with extensive experiments shows that blocking MP outperforms signed propagation under high edge error ratios, improving performance in both homophily and heterophily.
\end{abstract}

\section{Introduction}
Graph Neural Networks (GNNs) have shown remarkable performance with the aid of a message-passing (MP), where the representation of each node is recursively updated using its neighboring nodes based on structural properties \cite{defferrard2016convolutional,kipf2016semi,velickovic2017graph}. Early GNNs rely on network homophily, assuming that connected nodes will likely share similar labels. However, many real-world graphs have low homophily (a.k.a. heterophilic) \cite{zhu2020beyond,ma2021homophily,luan2022revisiting}, where spectral-based GNNs \cite{wang2022powerfula} achieve dismal performance under this condition since Laplacian smoothing only receives low-frequency signals from neighboring nodes.

To separate the embeddings of connected but dissimilar nodes, recent algorithms employ high-pass filters by adjusting edge weights during MP \cite{velickovic2017graph,brody2021attentive,chen2023lsgnn,liao2024ld2}. Notably, flipping the sign of edges from positive to negative known as signed propagation \cite{chien2020adaptive,bo2021beyond}, or blocking MP by assigning zero weights to heterophilic connections \cite{luo2021learning,hu2021graph,tian2022learning} has recently achieved remarkable performance. These techniques represent significant advancements in the field, offering new ways to enhance the performance and applicability of GNNs in diverse graph structures.

Recently, efforts have been made to theoretically analyze the effect of various propagation schemes in terms of node separability \cite{ma2021homophily,yan2021two,baranwal2023effects} building upon the insights of the Contextual Stochastic Block Model (CSBM) \cite{deshpande2018contextual}. In detail, they measure the distance of node feature expectations from a decision boundary, demonstrating that signed propagation outperforms plain message-passing algorithms in binary classification tasks. More recently, \cite{choi2023signed,li2024metadata} provided new insights by extending the analysis to multiple classes and considering degree distributions. However, few efforts have been made to determine whether signed MP consistently improves separability compared to not using propagation.

In this paper, we focus on the work of \cite{luo2021learning}, which mitigates local smoothing by assigning zero weights (blocking information) to heterophilic edges. Building on prior theoretical analysis, we propose the estimation of two latent parameters during training: homophily and edge classification error ratios. Based on this, we argue that MP (even with signed propagation) may result in poor performance compared to not propagating under specific conditions. Therefore, we suggest adaptively blocking information based on these values. Finally, since these values may not be easily measurable during the training phase, we propose an estimation strategy using the validation score. In summary, our contributions are as follows:
\begin{itemize}
    \item Unlike previous methods that focused solely on signed propagation to prevent over-smoothing, we demonstrate that message passing may degrade the separability of graph neural networks under certain conditions.
    \item To mitigate the smoothing effect, we propose an adaptive propagation approach based on the estimated parameters. We show that blocking information can be more efficient depending on the homophily and edge error ratios.
    \item Extensive experiments on real-world benchmark datasets with state-of-the-art baselines show notable performance improvements, validating the proposed scheme.
\end{itemize}


\section{Preliminaries} \label{preliminary}
\textbf{Notations.} Let $\mathcal{G}=(\mathcal{V},\mathcal{E}, X)$ be a graph with $\vert\mathcal{V}\vert=n$ nodes and $\vert\mathcal{E}\vert=m$ edges. The node attribute matrix is $X \in \mathds{R}^{n \times F}$, where $F$ is the dimension of an input vector. Given $X$, the hidden representation of node features $H^l$ at the \textit{l-th} layer is derived through message-passing. Here, node $i's$ feature is defined as $h_i^l$.
The structural property of $\mathcal{G}$ is represented by its adjacency matrix $A \in \{0, 1\}^{n \times n}$. A diagonal matrix with node degrees $D$ is derived from $A$ as $d_{ii}=\sum^n_{j=1}{A_{ij}}$.
Each node has its label $Y \in \mathds{R}^{n \times C}$ ($C$ represents the number of classes). 

\textbf{Message-Passing (MP).} \label{general_gnn}
Generally, GNNs employ alternate steps of propagation and aggregation recursively, during which the node features are updated iteratively. The widely known MP algorithm is GCN \cite{kipf2016semi}, which can be represented as:
\begin{equation}
\label{gnn}
\begin{gathered}
H^{l+1}=\sigma(\tilde{A}^lH^{l}W^{l})
\end{gathered}
\end{equation}
Here, $H^{0}=X$ is the initial vector and $H^{l}$ is nodes' hidden representations at the \textit{l-th} layer. $H^{l+1}$ is retrieved through message-passing ($\tilde{A}=D^{-1}A$) with an activation function $\sigma$. $W^{l}$ is trainable weight matrices. The final prediction is produced by applying cross-entropy $\sigma(\cdot)$ (e.g., log-softmax) to $H^L$ and the loss function is defined as below:
\begin{equation}
\label{loss_gnn}
\mathcal{L}_{GNN}=\mathcal{L}_{nll}(Y, \widehat{Y}) \leftarrow \,\,\widehat{Y}=\sigma(H^L)
\end{equation}
The loss is computed through a negative log-likelihood $\mathcal{L}_{nll}$ between true labels ($Y$) and predictions, $\widehat{Y}=\sigma(H^L)$.

\textbf{Homophily.} $\mathcal{H}_g$ stands for the global edge homophily ratio, which is defined as:
\begin{equation}
\label{global_homo}
\mathcal{H}_g \equiv \frac{\sum_{(i,j)\in \mathcal{E} 1(Y_i=Y_j)}}  {|\mathcal{E}|}
\end{equation}
Likewise, the local homophily ratio, $b_i$, of node $i$ is given as:
\begin{equation}
\label{local_homo}
b_i \equiv \frac{{\sum^n_{j=1} A_{ij} \cdot 1(Y_i=Y_j)}} {d_{ii}}
\end{equation}
Given a partially labeled training set $\mathcal{V}_L$, the goal of semi-supervised node classification is to correctly predict the classes of unlabeled nodes $\mathcal{V}_U=\{\mathcal{V}-\mathcal{V}_L\} \subset \mathcal{V}$. 

\textbf{Over-smoothing.} As introduced in \cite{oono2019graph,cai2020note}, over-smoothing measures the distinguishability of node features as follows:
\begin{equation}
\label{smoothing}
    \mu (H^l) := ||H^l - 1{\frac{1^TH^l}{N}}||_F
\end{equation} 
The over-smoothing happens if $\lim\limits_{l \rightarrow \infty} \mu (H^l) = 0$, where the node representation converges to zero after infinite propagation. Recently, \cite{mao2024demystifying} prove that even the attention-based GNN \cite{velickovic2017graph} loses the separability exponentially as $l$ increases.

\section{Motivation} 
We first define various MP schemes, including plain, signed, and blocking (pruning) types. Then, we introduce the prior theorems on local smoothing using the Contextual Stochastic Block Models. Finally, we highlight their drawbacks and present our new insights based on the estimated parameters.

\subsection{Local Smoothing on Message-Passing Schemes} \label{prior_thm}
We first define the concept of three MP schemes \cite{baranwal2021graph,yan2021two,choi2023signed} below.

\begin{manualdef}{1} [\textbf{M}essage-\textbf{P}assing Schemes]
    Building upon the Laplacian-based degree normalization \cite{kipf2016semi}, a.k.a. GCN, each propagation scheme modifies the adjacency matrix $\tilde{A}=D^{-1}A$ (Eq. \ref{gnn}) as follows:
    \begin{itemize}
    \item \textbf{Plane MP} 
    inherits the original matrix, which only consists of positive edges. 
    \begin{equation}
        \forall (i,j) \in \mathcal{E}, \,\,\, \tilde{A}=D^{-1}A \geq 0   
    \label{plane_mp}
    \end{equation}
    \item \textbf{Signed MP} 
    assigns negative values to the heterophilic edges, where $y_i \neq y_j$.
    \begin{equation}
        \forall (i,j) \in \mathcal{E}, \,\,\, \tilde{A} \in
    \begin{cases}
        D^{-1}A, & \,\,\, y_i = y_j\\
        -D^{-1}A, & \,\,\, y_i \neq y_j
    \end{cases}
    \label{signed_mp}
    \end{equation}
    \item \textbf{Blocked MP} 
    blocks the information propagation for heterophilic edges by assigning zero.
    \begin{equation}
        \forall (i,j) \in \mathcal{E}, \,\,\, \tilde{A} \in
    \begin{cases}
        D^{-1}A, & \,\,\, y_i = y_j\\
        0, & \,\,\, y_i \neq y_j
    \end{cases}
    \label{blocked_mp}
    \end{equation}
    \end{itemize}
\label{def_prop}
\end{manualdef}

To analyze the smoothing effect of each MP scheme, we inherit several useful notations defined in \cite{yan2021two} as follows: (1) For all nodes $i=\{1,...,n\}$, their degrees $\{d_i\}$ and latent features $\{h_i\}$ are \textit{i.i.d.} random variables. 
(2) Each class has the same population.
(3) The scale of each class distribution after initial embedding is identical, $||\mathds{E}(h^{(0)}_i|y_i)||=\mu$. Then, the feature distribution after a single hop propagation $\mathds{E}(h^{(1)} | y_i)$ can be defined through a \textbf{C}ontextual \textbf{S}tochastic \textbf{B}lock \textbf{M}odel (CSBM) as follows.
\begin{manualdef}{2}[CSBM under binary class scenario]
Assume a binary class $\mathds{E}(h^{(0)}_i|y_i) \sim (\mu, \theta=\{0,\pi\})$ and node feature is sampled from Gaussian distribution ($N$) \cite{deshpande2018contextual}. If $y_i=0$, the updated distribution $\mathds{E}(h^{(1)}|y_i)$ is given by:
\begin{equation}
\mathds{E}(h^{(1)}|y_i) \sim N({\mu},\frac{1}{\sqrt{\text{deg(i)}}})    
\label{csbm}
\end{equation}
\end{manualdef}
\cite{choi2023signed} extended binary CSBM (Eq. \ref{csbm}) to multiple (ternary) classes using additional angle $\phi$ as below:
\begin{equation}
\label{multi_expect}
\begin{gathered}
\mathds{E}(h^{(0)}_i|y_i)=(\mu, \phi,\theta),
\end{gathered}
\end{equation}
where $\phi=\pi / 2$ and $0 \leq \theta \leq 2\pi$.
Note that the above equation satisfies the binary case's origin symmetry as $(\mu,\pi/2,0)=-(\mu,\pi/2,\pi)$. Now, we define the impact of three MP schemes using a multi-class CSBM below.
\begin{manuallem}{3} [Multi-class CSBM, Plane MP] \label{multi_orig}
Given $y_i=0$, let us assume ego $k$ $\sim$ $(\mu,\pi/2,0)$ and aggregated neighbors $k'$ $\sim$ $(\mu,\pi/2,\theta')$. By replacing the $\mathds{E}_p(h^{(1)}_i|y_i,d_i)$ as $\mathds{E}_p(\cdot)$, the expectation after plane MP is as follows: 
\begin{equation}
\begin{gathered}
\mathds{E}_p(\cdot) 
= \frac{\{kb_i + k'(1-b_i)\}d'_i + k}{d_i+1}
\end{gathered}
\label{eq_plane}
\end{equation}
The $k'$ always satisfies $||k'|| \leq ||\mu||$ regardless of the normalized degree and homophily ratio since $1-b_i \leq 1$. 
\end{manuallem}

\begin{manuallem}{4} [Multi-class CSBM, Signed MP] \label{multi_flip}
Similar to above, the expectation $\mathbb{E}_s(\cdot)$ after signed MP is given by: 
\begin{flalign}
\label{eq_sign}
\mathbb{E}_s(\cdot) = \frac{(1-2e)\{b_ik+(b_i-1)k'\}d'_i+k}{d_i+1}
\end{flalign}
\end{manuallem}
\textbf{Remark.} The notation $e$ stands for the error ratio of incorrectly changing the sign of edges, which will be addressed throughout this paper. As introduced in \cite{choi2023signed}, the sign inconsistency caused by multi-hop propagation can be solved through jumping knowledge \cite{xu2018representation}.

\begin{manuallem}{5} [Multi-class CSBM, Blocked MP] \label{multi_obs}
Lastly, the expectation $\mathbb{E}_b(\cdot)$ driven by blocked MP can be defined as:
\begin{flalign}
\label{eq_block}
\mathbb{E}_b(\cdot) = \frac{\{(1-e)b_ik+e(1-b_i)k'\}d'_i+k}{d_i+1}
\end{flalign}
\end{manuallem}

Proof of Lemma \ref{multi_orig} - \ref{multi_obs} are in Appendix A.

\subsection{Necessity of Blocked MP}
Based on the analyses above, we derive new insights into the smoothing effect of each propagation scheme. Following \cite{yan2021two}, we assume that node separability (discriminative power) is proportional to the coefficient of $\mathds{E}_p(\cdot),\mathds{E}_s(\cdot),\mathds{E}_b(\cdot)$ in the above lemma. Through this assumption, we first show that plain MP has lower separability than signed and blocked MPs.
\begin{manualcoroll}{6} [Plane vs Other MPs] \label{comp1}
Since $d_i' / (d_i+1)$ is shared for all MPs (Eq. \ref{eq_plane} - \ref{eq_block}), we can compare the separability by omitting them as follows:
\begin{itemize}
    \item (Plane vs Signed MP) By comparing Eq. \ref{eq_plane} and \ref{eq_sign}: 
\begin{flalign}
Z_1 = \mathbb{E}_p(\cdot)-\mathbb{E}_s(\cdot) = (2e-1)\{b_ik+(b_i-1)k'\}
\end{flalign}
    \item (Plane vs Blocked MP) By comparing Eq. \ref{eq_plane} and \ref{eq_block}:
\begin{equation}
\begin{gathered}
Z_2 = \mathbb{E}_p(\cdot)-\mathbb{E}_b(\cdot) = eb_ik+(1-e)(1-b_i)k'
\end{gathered}
\end{equation}
\end{itemize}

The $\iint_{e,b_i} Z_1$ and $\iint_{e,b_i} Z_2$ are always negative, regardless of the variables $b_i$ and $e$, indicating that \textbf{plain MP has the lowest discrimination power among the three MPs}. Therefore, we can focus on the separability of the signed and blocked MPs to determine the optimal propagation scheme.
\end{manualcoroll}
\begin{manualcoroll}{7} [Signed vs Blocked MP] \label{coroll_2}
The difference between signed and blocked MP is given by:
\begin{equation}
\label{eq_disc}
Z_3=\mathbb{E}_s(\cdot)-\mathbb{E}_b(\cdot)=(1-2e)k+(b_i-e)k'
\end{equation}
\end{manualcoroll}
Previous studies assume a perfect edge classification scenario ($e=0$), leading to the conclusion that $Z=1-b_i \geq 0$, where signed MP outperforms blocked MP. However, achieving this optimal condition is challenging in a semi-supervised setting with few training nodes. Therefore, we propose estimating two parameters, $e$ and $b_i$, to select the most precise MP schemes. To facilitate understanding, we provide an example by varying $e$ below.
\begin{manualcoroll}{8} [Numerical example on edge error ratio] \label{coroll_3}
We assume three different edge error ratios, $e={1, 0.5, 0}$ to compare the signed and blocked MPs. Assuming that the neighbors are \textit{i.i.d.} ($k'=-k$), Eq. \ref{eq_disc} simplifies to:
\begin{equation}
\label{z3_updated}
    Z_3=(1-e-b_i)k=\begin{cases}
        -b_ik, \,\,\, &  \,\,\, e=1\\
        (-b_i+0.5)k, \,\,\, &  \,\,\, e=0.5\\
        (1-b_i)k, \,\,\, &  \,\,\, e=0\\
    \end{cases}
\end{equation}
\label{coroll_new_e}
\end{manualcoroll}
We can infer some useful insights from the above Corollary: (1) Under a high error ratio ($e=1$, initial stage of training), it might be better to \textit{not propagate} rather than using signed edges since $-b_ik \leq 0$ ($Z_3 \leq 0$). In addition, an error in signed propagation increases the uncertainty more than blocked GNNs as shown in Lemma \ref{lem_uncertainty}. (2) If the error is mediocre ($e=0.5$), the sign of $Z_3$ is dependent on the homophily ratio $b_i$. In this condition, signed MP may perform well ($Z_3 \geq 0$) under heterophily ($b < 0.5$), but it is still advantageous not to propagate messages under homophily. (3) However, if the edges are perfectly classified ($e=0$), signed MP might be the best option ($1-b_i \geq 0$), as demonstrated in prior work \cite{choi2023signed}. Therefore, we suggest the estimation of these two parameters, $b_i$ and $e$ for precise training. In addition, the following theorem reveals the necessity of blocked MP under high $e$ in terms of uncertainty.
\begin{manualtheorem}{9} [Uncertainty]
    Under large values of $e$, signed MP exhibits higher entropy than the blocked ones.

    Proof can be found in Appendix B.
\label{lem_uncertainty}
\end{manualtheorem}

\section{Methodology} \label{methodology}
Selecting an appropriate MP scheme is crucial for reducing smoothing and uncertainty. However, this may not be tractable in a semi-supervised learning context. Thus, we employ an EM algorithm, where the \textbf{E-step} is used for parameter estimation, and the \textbf{M-step} is used for optimization.

\subsection{(E-Step) Parameter Estimation}
We start with the strategy of homophily ($b_i$) and edge error ($e_t$) estimation in Theorem \ref{thm_homo} and \ref{thm_edg}, respectively. 
\begin{manualtheorem} {10} [Homophily estimation]
    The homophily ratio $b_i$ can be inferred using the mechanism of MLP \cite{wang2022powerfulb} and EvenNet \cite{lei2022evennet} as follows:
    \begin{equation}
    \label{homo_est}
    \begin{gathered}
        b_i = B_iB_i^T, \,\, B := \sigma\left(\sum^L_{l=0} XW^l + \sum^{\lfloor L/2 \rfloor}_{l=0}\tilde{A}^{2l}XW^l\right)
    \end{gathered}
    \end{equation}
which can take advantage of both the heterophily robustness of MLP and the low variance of EvenNet. 

The proof is provided in Appendix C. 
\label{thm_homo}
\end{manualtheorem}

\begin{figure}[t]
 \includegraphics[width=.48\textwidth]{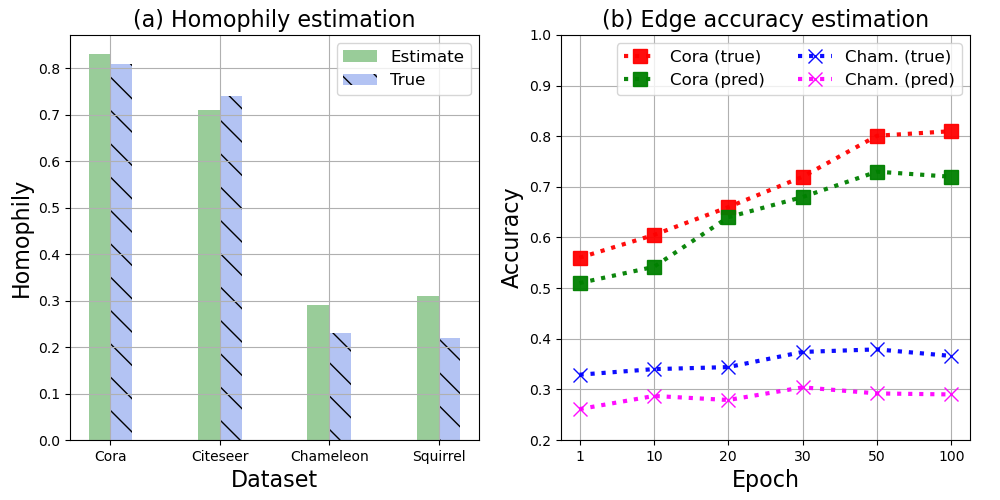}
    \caption{Empirical analysis on (a) homophily and (b) edge error estimation. More details can be found in Appendix D}
  \label{verify}
\end{figure}

\begin{manualtheorem}{11} [Edge error estimation]
    Given the node classification accuracy of validation sets ($\alpha_{t-1}$) at iteration $t-1$ and the number of classes ($c$), $e_t$ can be inferred as below:
    \begin{equation}
    \label{edge_err_est}
        e_t = 1 - \{a^2_{t-1}+\frac{(1-a_{t-1})^2}{c-1}\}
    \end{equation}
    \textbf{Approximation error.} Given the experimental settings in \cite{kipf2016semi}, the number of validation sets satisfies $n_{val} = 1,080 > 30$ for Cora as in Table \ref{dataset}. Thus, it can be easily inferred that the feature distribution of randomly sampled validation sets ($\mathbb{E}[h_{val}]$) follows the total distribution ($\mathbb{E}[h]$) as below. 
    \begin{equation}
        Z = \frac{\mathbb{E}[h_{val}] - \mathbb{E}[h]}{var(h_{val}) / \sqrt{n_{val}}} \sim N(0,1)
    \end{equation}
    Now, we demonstrate that the validation samples are less likely to be biased using the concentration theorem. Given that the output of the GNN ($h_i$) follows a Dirichlet distribution, Dir($h;c) = \frac{1}{B(c_1,...,c_K)}\prod^K_{k=1} h^{c_k-1}_i$, and assuming the classes are \textit{i.i.d.}, we can derive the following inequality:
    \begin{equation}
        \mathbb{P}(||h_{val}-\mathbb{E}[h_{val}]||_2 \geq \varepsilon) \leq 2\cdot \exp(-2\varepsilon^2 n_{val})
    \end{equation}
    which means that the validation nodes are equally distributed around its center $\mathbb{E}[h_{val}]$. 
    
    The derivation of Eq. \ref{edge_err_est} is provided in Appendix E. 
\label{thm_edg}
\end{manualtheorem}

\textbf{Empirical analysis.} In Figure \ref{verify}, we conduct empirical analyses to verify Theorem \ref{thm_homo} (left) and \ref{thm_edg} (right). In the figure (a), we illustrate the estimated and true homophily ratios under semi-supervised settings \cite{kipf2016semi}. In the figure (b), we compare the true and estimated edge errors (Eq. \ref{edge_err_est}) for two graphs, Cora (homophilic) and Chameleon (heterophilic), during the training epochs (x-axis). As shown in both figures, the predicted values are very similar to the actual values, with minor differences likely due to the node distribution of the benchmark graph not being \textit{i.i.d.} condition. Detailed experimental settings and additional results can be found in Appendix F, where the estimated values are almost identical if classes are identically distributed.

\begin{algorithm}[t]
\caption{Pseudo-code of our method}
\label{algo}
\begin{algorithmic}[1]
\Require Adjacency matrix ($\tilde{A}$), node features ($X$), initialized parameters ($\theta$), initialized validation ($\alpha_0=0$) and best validation score ($\alpha^*=0$)
\Ensure Parameters with the best validation score ($\theta^*$)
\State Homophily estimation $b_i$ (Eq. \ref{homo_est})
\For{training epochs $t \geq 1$}
\State \textbf{(E-Step)}
\State Get validation score $a_{t-1}$
\State Edge error estimation $e_t$ (Eq. \ref{edge_err_est})
\State \textbf{(M-Step)}
\State For all edges in $\tilde{A}$
\If {$\tilde{A}_{ij}<0 \,\, \wedge \,\, Z_t=1-b_i-e_t < 0$}
\State $\widehat{A}_{ij}=0$
\Else
\State $\widehat{A}_{ij}=\tilde{A}_{ij}$
\EndIf
\State Message-passing, $H^{(l+1)}=\sigma(\widehat{A}H^{(l)}W^{(l)})$
\State Node classification, $\mathcal{L}_{GNN}=\mathcal{L}_{nll}(Y, \sigma(H^{(L)}))$
\State Parameter update, $\theta^{(t+1)}=\theta^t-\eta{\partial\mathcal{L}_{GNN} / \partial\theta^t}$
\State Get validation score $\alpha_t$
\If {$\alpha_t$ $>$ $\alpha^*$}
\State Save current parameters as best, $\theta^*=\theta^t$
\State Update best validation score, $\alpha^*=\alpha_t$
\EndIf
\EndFor
\end{algorithmic}
\end{algorithm}

\subsection{(M-Step) Optimization with Calibration}
Referring to the separability of signed and blocked MP in Eq. \ref{z3_updated}, we can determine the propagation type based on the estimated values of $b_i$ (Eq. \ref{homo_est}) and $e$ (Eq. \ref{edge_err_est}). For each training step $t$, the discrimination gap between signed and blocked MP can be defined below: 
\begin{equation}
\label{threshold}
    Z_t=(1-b_i-e_t)k
\end{equation}
If $Z_t < 0$, we block signed messages to reduce the smoothing effect (Corollary \ref{coroll_2}) and uncertainty (Lemma \ref{lem_uncertainty}). For a numerical definition, let us assume the adjacency matrix derived from the downstream task as $\tilde{A}$ as defined from Eq. \ref{plane_mp} to \ref{blocked_mp}. Then, we calibrate the edges based on the following conditions:
\begin{equation}
    \widehat{A}_{ij} = \begin{cases}
        0, \,\,\, & \,\,\, \tilde{A}_{ij}<0 \,\, \wedge \,\, Z_t < 0 \\
        \tilde{A}_{ij}, \,\,\, & \,\,\, \text{otherwise}\\
    \end{cases}
\label{edge_calib}
\end{equation}
We replace the normalized adjacency matrix ($\tilde{A}$) in Eq. \ref{gnn} with $\widehat{A}$ as follows:
\begin{equation}
\label{new_msg}
H^{(l+1)}=\sigma(\widehat{A}H^{(l)}W^{(l)})
\end{equation}
Then, we employ Eq. \ref{loss_gnn} ($\mathcal{L}_{GNN}$) for optimization, where the above strategy can be applied to general signed MPs.

\subsection{Theoretical Justification} 
We aim to show that the proposed method can relieve the smoothing effect of signed MP using the notion of spectral radius (Def. \ref{def_spec_rad} to Thm. \ref{thm_substoc}) and separability (Thm. \ref{thm_disc}). 

\begin{manualdef}{12} [Spectral Radius] Let $\lambda_1,...,\lambda_n$ be the eigenvalues of an adjacency matrix $A \in \mathcal{R}^{n \times n}$. Then, the spectral radius of $A$ is given by:
\begin{equation}
\label{spec_rad}
    \rho(A) = \max \{|\lambda_1|,...,|\lambda_n|\}
\end{equation}
$A^{\infty}$ is well known to converge if $\rho(A)<1$ or $\rho(A)=1$, where $\lambda_1=1$ is the only eigenvalue on the unit circle.
\label{def_spec_rad}
\end{manualdef}

\begin{manualdef}{13} [Joint Spectral Radius] The generalization of a spectral radius
from a single matrix to a finite set of matrices $\mathcal{M}=\{A_1, ..., A_n\}$:
\begin{equation}
\label{jsr}
    J_{\rho}(A)=\lim_{k \rightarrow \infty}\max\{||A_1 \cdots A_k ||^{1/k}:A \in \mathcal{M}\}
\end{equation}
where the over-smoothing occurs at an exponential rate if and only if $J_{\rho}(A)<1$. In the following lemma, we first demonstrate that even attention-based MP cannot resolve the over-smoothing issue.
\end{manualdef} 
\begin{manuallem}{14} [Attention is susceptible to over-smoothing] 
For all layers $l$, the attention matrix $\tilde{A}^l$ cannot change the connectivity of the original graph \cite{brody2021attentive} as follows:
\begin{equation}
    0 < \tilde{A}^l_{ij} \leq 1, \,\,\, \forall (i,j) \in \mathcal{E}
\end{equation}
It is well known that the infinite products of non-homogeneous row-stochastic matrices $\Pi^{\infty}_{l=1}\tilde{A}^l$ converge to the same vector \cite{cowles1996markov,seneta2006non}, where we can easily infer that the stacking the infinite softmax-based attention matrix also converges (irreducible).
Recent work \cite{wu2024demystifying} provided a tighter bound that GAT satisfies $J_{\rho}(A)<1$ (Eq. \ref{jsr}), which means that $\lim\limits_{l \rightarrow \infty}\tilde{A}^l$ converges at an exponential rate.
\label{lem_conv}
\end{manuallem}

Unlike an attention-based matrix, where the total sum of the edge weights connected to a neighbor node is 1, a signed adjacency matrix (Eq. \ref{signed_mp}) is sub-stochastic since the sum of rows is less than 1. Therefore, the spectral radius of the sub-stochastic matrix satisfies the following conditions.
\begin{manualcoroll}{15} [Convergence of signed MP]
Let us assume that signed MP makes the original adjacency matrix $A$ as an irreducible sub-stochastic matrix $\tilde{A}$ (Eq. \ref{signed_mp}). Before the proof, note that the corresponding eigenvector of $\rho(\bar{A})$ satisfies $||\lambda||_1=1$ and $\epsilon_i=1-\sum^N_{j=1}A_{ij}$. Referring to Eq. \ref{spec_rad}, the spectral radius of the sub-stochastic matrix satisfies the condition $\rho(\bar{A})<1$ as below:
\begin{equation}
\rho(\bar{A})=\underset{\text{stochastic matrix, }\rho(A)=1}{\underline{\sum_{i=1}\sum_{j=1}\lambda_i(A_{ij}+\epsilon_{ij})}}-\underset{>0}{\underline{\sum_{i=1}\sum_{j=1}\epsilon_{ij}}} < 1\\
\end{equation}
which means signed MP also converges. However, our edge calibration can mitigate the smoothing effect as follows. 
\label{rad_substo}
\end{manualcoroll}
\begin{manualtheorem}{16}[Edge calibration reduces local smoothing]
Based on Corollary \ref{rad_substo}, we can infer that a matrix with k-disconnected sub-stochastic components satisfies $|\lambda_1|=|\lambda_2|=\cdots=|\lambda_k|=1$. According to Definition \ref{def_spec_rad}, $J_{\rho}(\bar{A})=1$ with multiple $\lambda$'s on the unit circle does not meet the convergence property compared to the graph attention as in Lemma \ref{lem_conv}, relieving the smoothing effect in Eq. \ref{smoothing}.
\label{thm_substoc}
\end{manualtheorem}

Lastly, we prove that edge calibration also enhances the separability of signed MP as follows.
\begin{manualtheorem}{17} [Separability]
Signed MP with edge calibration outperforms the original signed and blocked MPs.
\begin{itemize}
    \item (High $e_t \geq 0.5$, Ours (blocked) vs Signed MPs) \begin{equation}
        Z_4=\mathbb{E}_b(\cdot)-\mathbb{E}_s(\cdot)=\int_{b=0}^{1}\int_{e_t=0.5}^{1} (1-b_i-e_t+2e_tb_i) =\frac{9}{8}
    \end{equation}
    \item (Low $e_t < 0.5$, Ours (signed) vs Blocked MPs)
    \begin{equation}
        Z_5=\mathbb{E}_s(\cdot)-\mathbb{E}_b(\cdot)=\int_{b=0}^{1}\int_{e_t=0}^{0.5} (1-e_t-b_i) = \frac{1}{8}
    \end{equation}
\end{itemize}
\label{thm_disc}
Both equations satisfy $\iint_{e,b_i} Z_4 > 0$ and $\iint_{e,b_i} Z_5 > 0$, indicating that edge calibration leverages the advantages of both signed and blocked MPs under each condition.
\end{manualtheorem}


\begin{table}[t]
\caption{Details of the six benchmark graphs. We follow the experimental settings of GCN \cite{kipf2016semi}}
\label{dataset}
\centering
\begin{adjustbox}{width=0.48\textwidth}
\begin{tabular}{@{}llllllll}
&     &        &         &  & & & \\ 
\Xhline{2\arrayrulewidth}
        & Datasets         & Cora  & Citeseer & Pubmed & Actor & Cham. & Squirrel \\ 
\Xhline{2\arrayrulewidth}
                        & \# Nodes  & 2,708  & 3,327   & 19,717 & 7,600 & 2,277  & 5,201 \\
                        & \# Edges         & 10,558  & 9,104  & 88,648   & 25,944 & 33,824  & 211,872 \\
                        & \# Features       & 1,433  & 3,703  & 500   & 931 & 2,325  & 2,089 \\
                        & \# Labels        & 7  & 6  & 3     & 5  & 5  & 5 \\
                        & \# Training & 140 & 120 & 60 & 100 & 100 & 100 \\
                        & \# Validation & 1,083 & 1,330 & 7,886 & 3,040 & 910 & 2,080 \\
\Xhline{2\arrayrulewidth}
\end{tabular}
\end{adjustbox}
\end{table}

\begin{table*}[ht]
\caption{(Q1) Node classification performance (\%) with standard deviation ($\pm$) on the six benchmark graphs. The gray-colored cells indicate top-3 performance. A symbol $*$ means that edge calibration (Eq. \ref{edge_calib}) is applied on a base method} 
\label{perf_1}
\centering
\begin{center}
\begin{adjustbox}{width=.97\textwidth}
\begin{tabular}{@{}llllllll}
\Xhline{2\arrayrulewidth}
        & Datasets                       & Cora & Citeseer & Pubmed & Actor & Chameleon & Squirrel \\ 
        & $\mathcal{H}_g$ (Eq. \ref{global_homo}) & 0.81 & 0.74 & 0.8 & 0.22 & 0.23 & 0.22 \\
\Xhline{2\arrayrulewidth}
                        & GCN \cite{kipf2016semi}              & 80.5 ${\,\pm\,0.73 }$   & 70.4 ${\,\pm\,0.62 }$  & 78.6 ${\,\pm\,0.44 }$  & 20.2 ${\,\pm\,0.40 }$   & 49.3 ${\,\pm\,0.58 }$  & 30.7 ${\,\pm\,0.70 }$   \\
                        & GAT \cite{velickovic2017graph}            & 81.2 ${\,\pm\,0.51 }$  & 71.3 ${\,\pm\,0.75 }$  & 79.0 ${\,\pm\,0.45 }$  & 22.5 ${\,\pm\,0.36 }$  & 48.8 ${\,\pm\,0.83 }$  & 30.8 ${\,\pm\,0.94 }$  \\
                        & APPNP \cite{gasteiger2018predict}                & 81.8 ${\,\pm\,0.51 }$  & 71.9 ${\,\pm\,0.38 }$  & 79.0 ${\,\pm\,0.30 }$  & 23.8 ${\,\pm\,0.32 }$  & 48.0 ${\,\pm\,0.77 }$  & 30.4 ${\,\pm\,0.61 }$  \\
                        & GCNII \cite{chen2020simple}          & 81.3 ${\,\pm\,0.69 }$  & 70.7 ${\,\pm\,1.28 }$  & 78.5 ${\,\pm\,0.50 }$  & 25.9 ${\,\pm\,1.21 }$  & 48.6 ${\,\pm\,0.76 }$  & 30.4 ${\,\pm\,0.90 }$  \\
                        & H$_2$GCN \cite{zhu2020beyond}       & 79.4 ${\,\pm\,0.45 }$  & 71.2 ${\,\pm\,0.79 }$  & 78.1 ${\,\pm\,0.31 }$  & 25.6 ${\,\pm\,1.15 }$  & 47.5 ${\,\pm\,0.82 }$  & 31.0 ${\,\pm\,0.74 }$  \\
                        & PTDNet \cite{luo2021learning}         & 81.1 ${\,\pm\,0.79 }$  & 71.4 ${\,\pm\,1.12 }$  & 78.8 ${\,\pm\,0.67 }$  & 21.5 ${\,\pm\,0.75 }$  & 50.4 ${\,\pm\,1.06 }$  & 32.2 ${\,\pm\,0.75 }$  \\
                        & P-reg \cite{yang2021rethinking}      & 81.0 ${\,\pm\,0.88 }$ & 71.9 ${\,\pm\,0.81 }$ & 78.5 ${\,\pm\,0.42 }$ & 21.2 ${\,\pm\,0.52 }$ & 50.6 ${\,\pm\,0.37 }$ & \cellcolor{Gray}\textbf{33.1} ${\,\pm\,0.40 }$ \\
                        & ACM-GCN \cite{luan2022revisiting} & 81.6 ${\,\pm\,0.85 }$ & 71.3 ${\,\pm\,1.01 }$ & 78.4 ${\,\pm\,0.53 }$ & 24.9 ${\,\pm\,2.17 }$ & 49.6 ${\,\pm\,0.59 }$ & 31.2 ${\,\pm\,0.44 }$ \\
                        & HOG-GCN \cite{wang2022powerfulb} & 81.7 ${\,\pm\,0.41 }$ & 72.2 ${\,\pm\,0.67 }$ & 79.0 ${\,\pm\,0.24 }$ & 21.3 ${\,\pm\,0.56 }$ & 47.9 ${\,\pm\,0.45 }$ & 30.2 ${\,\pm\,0.50 }$ \\
                        & JacobiConv \cite{wang2022powerfula} & 81.9 ${\,\pm\,0.69 }$ & 72.0 ${\,\pm\,0.75 }$ & 78.7 ${\,\pm\,0.42 }$ & 26.0 ${\,\pm\,1.04 }$ & 51.6 ${\,\pm\,1.10 }$ & 32.1 ${\,\pm\,0.73 }$ \\
                        & GloGNN \cite{li2022finding} & 82.0 ${\,\pm\,0.40 }$ & 71.8 ${\,\pm\,0.52 }$ & 79.4 ${\,\pm\,0.28 }$ & \cellcolor{Gray}\textbf{26.6} ${\,\pm\,0.71 }$ & 48.3 ${\,\pm\,0.39 }$ & 30.8 ${\,\pm\,0.80 }$ \\
                        & AERO-GNN \cite{lee2023towards} & 81.6 ${\,\pm\,0.54 }$ & 71.1 ${\,\pm\,0.62 }$ & 79.1 ${\,\pm\,0.47 }$ & 25.5 ${\,\pm\,1.08 }$ & 49.8 ${\,\pm\,2.33 }$ & 29.9 ${\,\pm\,1.96 }$ \\
                        & Auto-HeG \cite{zheng2023auto} & 81.5 ${\,\pm\,1.06 }$ & 70.9 ${\,\pm\,1.41 }$ & \cellcolor{Gray}\textbf{79.2} ${\,\pm\,0.24 }$ & 26.1 ${\,\pm\,0.98 }$ & 48.7 ${\,\pm\,1.37 }$ & 31.5 ${\,\pm\,1.11 }$ \\
                        & TED-GCN \cite{yan2024trainable} & 81.8 ${\,\pm\,0.88 }$ & 71.4 ${\,\pm\,0.56 }$ & 78.6 ${\,\pm\,0.30 }$ & 26.0 ${\,\pm\,0.95 }$ & 50.4 ${\,\pm\,1.21 }$ & \cellcolor{Gray}\textbf{33.0} ${\,\pm\,0.98 }$ \\
                        & PCNet \cite{li2024pc} & 81.5 ${\,\pm\,0.76 }$ & 71.2 ${\,\pm\,1.20 }$ & 78.8 ${\,\pm\,0.26 }$ & 26.4 ${\,\pm\,0.85 }$ & 48.1 ${\,\pm\,1.69 }$ & 31.4 ${\,\pm\,0.56 }$ \\
\Xhline{2\arrayrulewidth}
                        & GPRGNN \cite{chien2020adaptive} & 81.1 ${\,\pm\,0.56 }$  & 71.0 ${\,\pm\,0.83 }$  & 78.7 ${\,\pm\,0.55 }$  & 24.8 ${\,\pm\,0.87 }$  & 50.2 ${\,\pm\,0.79 }$  & 30.2 ${\,\pm\,0.62 }$  \\
                        & \textbf{GPRGNN}$^*$ & \cellcolor{Gray}\textbf{82.5 ${\,\pm\,0.37 }$}  & \cellcolor{Gray}\textbf{72.4} ${\,\pm\,0.66 }$  & \cellcolor{Gray}\textbf{79.3} ${\,\pm\,0.35 }$  & \cellcolor{Gray}\textbf{26.9} ${\,\pm\,0.74 }$  & \cellcolor{Gray}\textbf{52.5} ${\,\pm\,0.54 }$  & 32.2 ${\,\pm\,0.49 }$  \\
                        & Relative improv. (+ \%) & + 1.73 \% & + 1.97 \% & + 0.76 \% & + 8.47 \% & + 4.58 \% & + 6.62 \% \\
\hdashline
                        & FAGCN \cite{bo2021beyond}       & 81.4 ${\,\pm\,0.51 }$  & 72.2 ${\,\pm\,0.69 }$  & 78.9 ${\,\pm\,0.62 }$  & 25.3 ${\,\pm\,0.77 }$  & 49.1 ${\,\pm\,1.20 }$  & 30.3 ${\,\pm\,0.96 }$  \\
                        & \textbf{FAGCN}$^*$ & \cellcolor{Gray}\textbf{82.8} ${\,\pm\,0.45 }$  & \cellcolor{Gray}\textbf{73.4 ${\,\pm\,0.50 }$}  & \cellcolor{Gray}\textbf{79.6} ${\,\pm\,0.33 }$  & \cellcolor{Gray}\textbf{27.8} ${\,\pm\,0.58 }$  & \cellcolor{Gray}\textbf{51.8} ${\,\pm\,0.91 }$  & 32.5 ${\,\pm\,0.77 }$  \\
                        & Relative improv. (+ \%) & + 1.72 \% & + 1.66 \% & + 0.89 \% & + 9.88 \% & + 5.50 \% & + 7.26 \% \\
\hdashline
                        & GGCN \cite{yan2021two} & 81.2 ${\,\pm\,1.06 }$  & 71.5 ${\,\pm\,1.44 }$  & 78.3 ${\,\pm\,0.35 }$  & 23.7 ${\,\pm\,0.75 }$  & 50.0 ${\,\pm\,0.98 }$  & 30.4 ${\,\pm\,0.72 }$  \\
                        & \textbf{GGCN}$^*$ & \cellcolor{Gray}\textbf{82.4} ${\,\pm\,0.87 }$  & \cellcolor{Gray}\textbf{73.0} ${\,\pm\,0.72 }$  & 79.0 ${\,\pm\,0.34 }$  & 25.6 ${\,\pm\,0.61 }$  & \cellcolor{Gray}\textbf{52.0} ${\,\pm\,0.72 }$  & \cellcolor{Gray}\textbf{32.7} ${\,\pm\,0.69 }$  \\
                        & Relative improv. (+ \%) & + 1.48 \% & + 2.10 \% & + 0.89 \% & + 8.02 \% & + 4.00 \% & + 7.57 \% \\
\Xhline{2\arrayrulewidth}
\end{tabular}
\end{adjustbox}
\end{center}
\end{table*}

\section{Experiments} \label{experiments}
We conduct extensive experiments to answer the following research questions:
\begin{itemize}
\item \textbf{Q1:} Does the proposed method improve the node classification accuracy of signed GNNs?
\item \textbf{Q2:} Does edge calibration enhance the inter-class separability of signed propagation?
\item \textbf{Q3:} How does performance change given the values $Z_t$ in Eq. \ref{threshold}?
\item \textbf{Q4:} How much does the method improve the quality of GNNs on large benchmark graphs?
\end{itemize}

\textbf{Baselines.} For the experiment, we classified the baselines into two main categories: signed GNNs and others. As shown in Table \ref{perf_1}, we applied our method to the following three signed propagation techniques for evaluation: GPRGNN \cite{chien2020adaptive}, FAGCN \cite{bo2021beyond}, and GGCN \cite{yan2021two}.

\textit{Please refer to Appendix G for the details of datasets, baselines, and implementations}

\subsection{Performance Analysis (Q1)} \label{ex_result}
Table \ref{perf_1} shows the node classification accuracy (\%) of several state-of-the-art methods. We analyze the results from the two perspectives below.

\textbf{Signed and blocked MPs Outperform Plane MP:} The experimental results demonstrate that GNNs with signed or blocked MP outperform traditional methods under benchmark graphs. For both homophilic and heterophilic datasets, we can see that plane MP-based algorithms like GCN, GAT, and APPNP show inferior accuracy than others. Especially, the performance gap increases as the homophily decreases. Specifically, on the Actor dataset, GloGNN relatively outperforms GCN by over 30\%. In addition, under Chameleon and Squirrel with many cyclic edges, Auto-HeG, signed (GPRGNN, FAGCN, and GGCN) or blocked (PTDNet) MPs achieve state-of-the-art performance. This highlights the efficacy of signed and blocked GNNs in capturing the complex relationships present in general graphs, thereby leading to superior improvement.

\textbf{Edge Weight Calibration Enhances the Performance of Signed GNNs:} The results underscore the critical importance of edge weight calibration in enhancing the performance of signed GNNs. Models that incorporate edge weight adjustments, such as GPRGNN$^*$ and FAGCN$^*$, consistently outperform their counterparts across a wide range of datasets, highlighting the robustness of this approach. For instance, GPRGNN$^*$ demonstrates remarkable performance improvements on both the Cora and Pubmed datasets compared to the baseline GPRGNN, indicating its adaptability and effectiveness. Similarly, FAGCN$^*$, including edge weight calibration, outperforms the original FAGCN model on homophilic and heterophilic graphs, showcasing its versatility in different graph structures. Additionally, our method leverages blocked MP by effectively removing cyclic edges for the Chameleon and Squirrel datasets, further enhancing model performance. These findings collectively suggest that edge weight calibration is crucial for maximizing the full potential of signed GNNs, as it plays a pivotal role in reducing the smoothing effect under edge uncertainty and ultimately achieving higher classification accuracy across diverse scenarios.

\begin{figure}[t]
 \includegraphics[width=.48\textwidth]{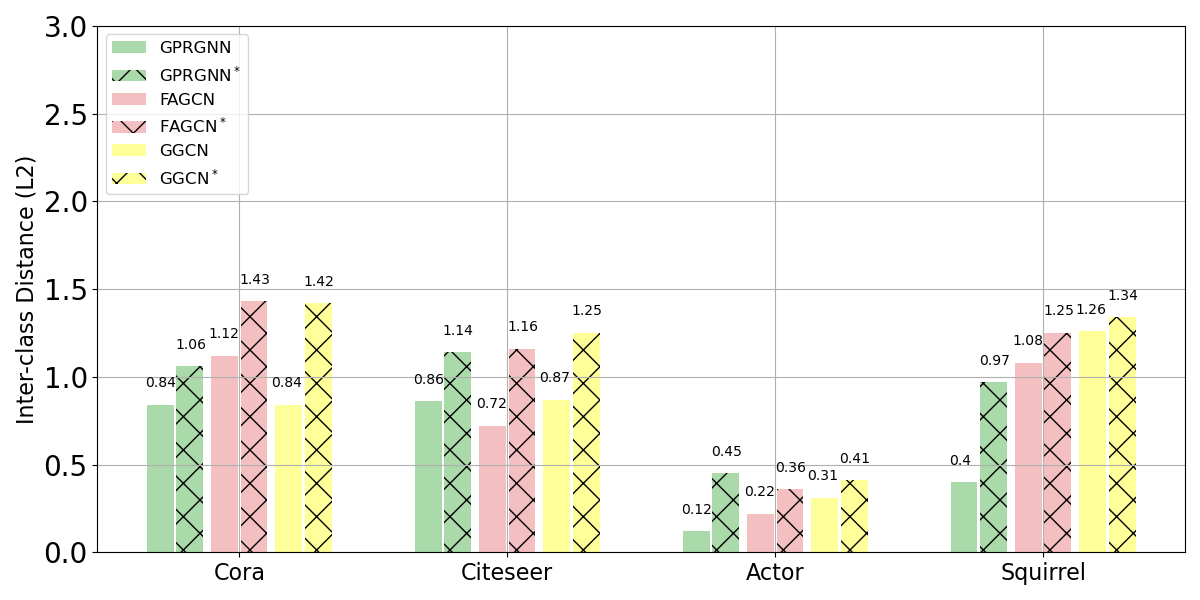}
    \caption{(Q2) We take three signed GNNs (GPRGNN, FAGCN, and GGCN) and measure the inter-class distances to show that our method improves the discrimination power}
  \label{q2}
\end{figure}

\subsection{Analysis on Discrimination Power (Q2)}
In Figure \ref{q2}, we present the inter-class distances for three GNNs with signed MP (GPRGNN, FAGCN, and GGCN) across four benchmark graphs (Cora, Citeseer, Actor, and Squirrel) to investigate a neural collapse perspective \cite{kothapalli2024neural}. The y-axis measures the average L2 distance between classes after the first layer projection. To ensure fairness, we removed parameter randomness from all baselines. The ensemble methods (indicated with an asterisk) consistently show an increase in inter-class distance compared to their standard counterparts across all datasets. For instance, on the Cora dataset, GPRGNN* and FAGCN* show increases to 1.06 and 1.43, respectively, compared to 0.84 and 1.12 for the standard models. The increase is even more substantial in heterophilic datasets like Actor and Squirrel, where the methods with edge calibration significantly improve separability compared to the standard versions, highlighting the effectiveness of blocked MP under certain conditions.

\begin{figure}[ht]
 \includegraphics[width=.48\textwidth]{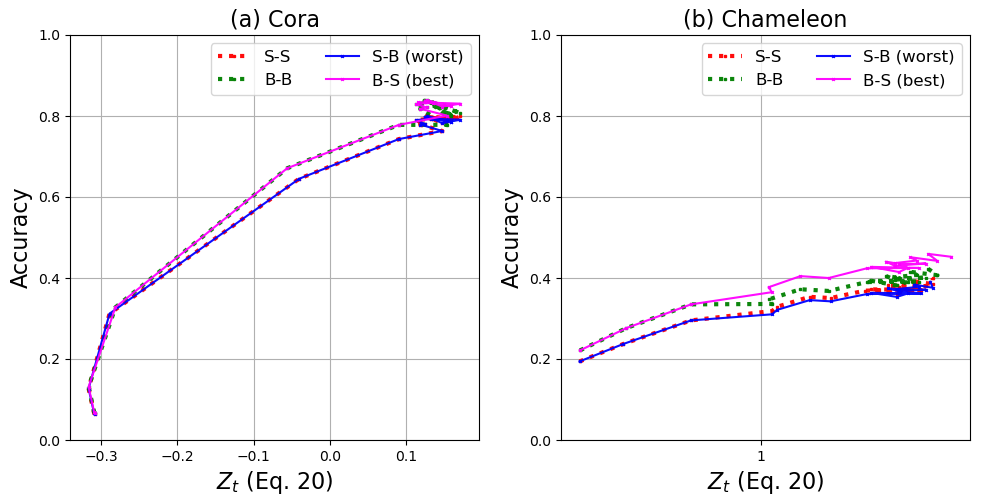}
    \caption{(Q3) Performance gain concerning signed/blocked propagation using FAGCN$^*$ model }
  \label{q3}
\end{figure}

\subsection{Importance of Proper MP Schemes (Q3)} \label{sec_ablation}
We aim to show that selecting a proper propagation scheme between signed or blocked MPs is crucial for GNNs. As described in Figure \ref{q3}, the x-axis stands for the $ Z_t $ in Eq. \ref{threshold}, and the y-axis represents node classification accuracy. For a fair comparison, we remove the randomness for all methods (e.g., parameter initialization) by fixing the seed. The left figure is (a) Cora and the right one is (b) Chameleon. We have proved that blocked MP outperforms signed MP in the case of $ Z_t < 0 $, and vice versa. To verify this, we assume two types of MPs: signed (S) and blocked (B) under two different conditions: $ Z_t < 0 $ and $ Z_t \geq 0 $. For example, S-S means that signed MPs are used independent of $ Z_t $, while B-S utilizes blocked MP if $ Z_t < 0 $ and signed MP for $ Z_t \geq 0 $. As illustrated, selecting the proper propagation scheme (B-S) achieves the best quality, where the performance gap becomes greater in the heterophilic graph (Chameleon). Conversely, improper MPs (S-B or S-S) when $ Z_t < 0 $ show low accuracy, showing that blocked MP can improve the quality of prediction under this condition significantly.

\begin{table}[ht]
\caption{(Q4) Node classification accuracy (\%) on large heterophilic graphs. Penn94 has a binary class, while the arXiv-year and snap-patents have five classes}
\centering
\begin{adjustbox}{width=0.48\textwidth}
\begin{tabular}{@{}lllll}
& \multicolumn{1}{l}{} &     &        &       \\ 
\Xhline{2\arrayrulewidth}
        & Datasets        & Penn94 & arXiv-year  & snap-patents \\ 
        & $\mathcal{H}_g$ (Eq. \ref{global_homo}) & 0.046 & 0.272 & 0.1 \\
\Xhline{2\arrayrulewidth}
                        & GCN \cite{kipf2016semi}  & 81.3 ${\,\pm\,0.3 }$  & 44.5 ${\,\pm\,0.3 }$  & 43.9 ${\,\pm\,0.2 }$ \\
                        & GAT \cite{velickovic2017graph}  & 80.6 ${\,\pm\,0.6 }$  & 45.0 ${\,\pm\,0.5 }$  & 45.2 ${\,\pm\,0.6 }$ \\
                        & GCNII \cite{chen2020simple}  & 81.8 ${\,\pm\,0.7 }$  & 46.1 ${\,\pm\,0.3 }$ & \textbf{47.5} ${\,\pm\,0.7 }$ \\
                        & H$_2$GCN \cite{zhu2020beyond}  & 80.4 ${\,\pm\,0.6 }$  & \textbf{47.6} ${\,\pm\,0.2 }$  & OOM \\
\Xhline{2\arrayrulewidth}
                        & GPRGNN \cite{chien2020adaptive}  & 80.9 ${\,\pm\,0.3 }$  & 43.6 ${\,\pm\,0.4 }$  & 41.4 ${\,\pm\,0.1 }$ \\
                        & \textbf{GPRGNN}$^*$       & \textbf{83.1} ${\,\pm\,0.4 }$ & 44.9 ${\,\pm\,0.3 }$  & 43.6 ${\,\pm\,0.3 }$ \\
                        & Relative improv. (+ \%) & 2.72\% & 2.98\% & 5.3\% \\
\Xhline{2\arrayrulewidth}
\end{tabular}
\end{adjustbox}
\label{large_dataset}
\end{table}

\subsection{Analysis on Large Graphs (Q4)}
We conduct experiments on large benchmarks \cite{lim2021large} and describe the results in Table 3. Due to the OOM issue, we apply our method to one of the memory-efficient signed GNNs, GPRGNN \cite{chien2020adaptive}. As shown in the table, our method GPRGNN$^*$ demonstrates notable improvements in node classification accuracy across large graphs. Specifically, on the three datasets, GPRGNN$^*$ achieves an accuracy improvement of 2.72\%, 2.98\%, and 5.3\% over the baseline GPRGNN, respectively. Although H$_2$GCN \cite{zhu2020beyond} and GCNII \cite{chen2020simple} show the best performance on arXiv and snap, this is because the accuracy of the original GPRGNN is quite lower compared to the other methods. To summarize, these results highlight the effectiveness of our approach in enhancing the performance of GNNs on large-scale benchmarks.

\section{Related Work}
\textbf{Heterophilic GNNs.} Starting from Laplacian decomposition \cite{defferrard2016convolutional}, spectral GNNs \cite{kipf2016semi,wang2022powerfula,du2022gbk,bo2023specformer,lu2024representation} have achieved remarkable performance on homophilic graphs. However, as the homophily \cite{platonov2024characterizing} of the graph decreases, their performance sharply declines due to local smoothing \cite{pei2020geom}. To address this limitation, spatial-based GNNs have emerged, developing many powerful schemes that adjust edge weights for message-passing \cite{velickovic2017graph,brody2021attentive,chen2023lsgnn,liao2024ld2}. Specifically, some studies handle disassortative edges by capturing node differences or incorporating similar remote nodes as neighbors \cite{derr2018signed,huang2019signed,zhu2020beyond,choi2022finding,lei2022evennet,wang2022powerfulb,zhao2023graph,zheng2023auto,mao2024demystifying,yan2024trainable,li2024pc,qiu2024refining,tang2024simcalib}. Among these, methods that either change the sign of the edge \cite{chien2020adaptive,bo2021beyond,fang2022polarized,guo2022clenshaw} or opt not to transmit information \cite{luo2021learning,hu2021graph,tian2022learning} have recently been proposed.

\textbf{Over-smoothing in GNNs.} In addition to the above methods, theoretical analyses have emerged explaining why each message-passing technique works well from the perspective of node separability (reduced smoothing effect). For example, \cite{yan2021two,baranwal2021graph} analyzed the separability of the plane, signed, and blocked propagation after message-passing under binary class graphs. Recently, \cite{choi2023signed} extended the theorems to multi-class scenarios, and \cite{li2024metadata} suggested their degree-corrected version. However, these methods assume a fixed edge classification error, which may fail to induce the smoothing effect precisely.

\section{Conclusion}
This paper presents a comprehensive study on the impact of edge uncertainty and over-smoothing in signed Graph Neural Networks (GNNs). First, we scrutinize the smoothing effect under different values of $Z_t$ (Eq. \ref{threshold}), offering a novel perspective that propagation schemes should account for varying homophily and edge classification error ratios. Second, we introduce an innovative training mechanism that dynamically selects between blocked and signed propagation based on these parameters, effectively mitigating over-smoothing and enhancing performance. Our theoretical analysis, supported by extensive experiments, demonstrates that blocking message propagation can be more effective than traditional message-passing schemes under certain conditions. This insight is crucial for improving node classification accuracy in both homophilic and heterophilic graphs. We hope that future work will explore further refinements of these techniques for more complex graph structures.



\bibliography{aaai24}

\bigskip


\section*{Technical Appendix}

\begin{appendices}
$\\$

\subsection{Appendix A (proof of lemma 3 - 5)}
In this section, we derive the expected distribution after three types of message-passing (MP) using a multi-class contextual stochastic block model (CSBM), respectively. Before delving into this, let us define these MP schemes again below.
\begin{itemize}
    \item \textbf{Plane MP} 
    inherits the original matrix, which only consists of positive edges. 
    \begin{equation}
        \forall (i,j) \in \mathcal{E}, \,\,\, \tilde{A}=D^{-1}A \geq 0
    \end{equation}
    \item \textbf{Signed MP} 
    assigns negative values to the heterophilic edges, where $y_i \neq y_j$.
    \begin{equation}
        \forall (i,j) \in \mathcal{E}, \,\,\, \tilde{A} \in
    \begin{cases}
        D^{-1}A, & \,\,\, y_i = y_j\\
        -D^{-1}A, & \,\,\, y_i \neq y_j
    \end{cases}
    \end{equation}
    \item \textbf{Blocked MP}  
    blocks the information propagation for heterophilic edges by assigning zero.
    \begin{equation}
        \forall (i,j) \in \mathcal{E}, \,\,\, \tilde{A} \in
    \begin{cases}
        D^{-1}A, & \,\,\, y_i = y_j\\
        0, & \,\,\, y_i \neq y_j
    \end{cases}
    \end{equation}
    \end{itemize}
$\\$

\textbf{A.1. Proof of Lemma 3 (Plane MP)}

Let's assume that $y_i=0$, ego $k$ $\sim$ $(\mu,\pi/2,0)$, and aggregated neighbors $k'$ $\sim$ $(\mu,\pi/2,\theta')$. Though each neighbor has multiple distributions proportional to the number of classes, their aggregation always satisfies $|k'| \leq \mu$ since the summation of coefficients ($1-b_i$) is lower than 1. Thus, we indicate $k'_{aggr}$ as $k'$ here for brevity. Given this, the expectation after plane MP $\mathds{E}_p(h^{(1)}_i|y_i,d_i)$ can be retrieved as below: 


\begin{flalign}
\mathds{E}_p(h^{(1)}_i|y_i,d_i) &={\frac{k}{d_i+1}} + \sum_{j \in \mathcal{N}_i}\left(\frac{k(1-e) + k e}{\sqrt{(d_i+1)(d_j+1)}}b_i+\frac{k'(1-e)+k' e}{\sqrt{(d_i+1)(d_j+1)}}(1-b_i)\right) \\
&=\frac{k}{d_i+1} + \sum_{j \in \mathcal{N}_i}\left(\frac{kb_i + k'(1-b_i) }{\sqrt{(d_i+1)(d_j+1)}}\right) \\
&= \frac{k}{d_i+1} + \frac{\{kb_i + k'(1-b_i)\}d'_i}{d_i+1} \\
&= \frac{\{kb_i + k'(1-b_i)\}d'_i + k}{d_i+1} \\
\end{flalign}

$\\$

\textbf{A.2. Proof of Lemma 4 (Signed MP)}

Similar to the above analysis, we can retrieve the expectation $\mathds{E}_s(h^{(1)}_i|y_i,d_i)$ after signed MP. For this, we employ the edge classification error rate $e$, which determines the error ratio of finding heterophilic edges as follows: 


\begin{flalign}
\mathds{E}_s(h^{(1)}_i|y_i,d_i)&=\frac{k}{d_i+1} + \sum_{j \in \mathcal{N}_i}\left(\frac{k(1-e) - k e}{\sqrt{(d_i+1)(d_j+1)}}b_i+\frac{-k'(1-e)+k' e}{\sqrt{(d_i+1)(d_j+1)}}(1-b_i)\right) \\
&= \frac{k}{d_i+1} + \sum_{j \in \mathcal{N}_i}\left(\frac{k(1-2e)b_i - k'(1-2e)(1-b_i)}{\sqrt{(d_i+1)(d_j+1)}}\right) \\
&= \frac{k}{d_i+1} + \sum_{j \in \mathcal{N}_i}\left(\frac{(1-2e)\{kb_i + k'(b_i-1)\}}{\sqrt{(d_i+1)(d_j+1)}}\right) \\
&= \frac{k }{ d_i+1} + \frac{(1-2e)\{kb_i + k'(b_i-1)\}d'_i}{d_i+1} \\
&= \frac{(1-2e)\{b_ik+(b_i-1)k'\}d'_i+k}{d_i+1}.
\end{flalign}

$\\$

\textbf{A.3. Proof of Lemma 5 (Blocked MP)}

Lastly, the expectation $\mathds{E}_b(h^{(1)}_i|y_i,d_i)$ of assigning zero weights for heterophilic edges is given by:

\begin{flalign}
\mathds{E}_b(h^{(1)}_i|y_i,d_i)&=\frac{k }{ d_i+1} + \sum_{j \in \mathcal{N}_i}\left(\frac{k(1-e)  - k e \times 0 }{ \sqrt{(d_i+1)(d_j+1)}}b_i+\frac{-k'(1-e)\times 0+k' e  }{ \sqrt{(d_i+1)(d_j+1)}}(1-b_i)\right) \\
&= \frac{k }{ d_i+1} + \sum_{j \in \mathcal{N}_i}\left(\frac{k(1-e)b_i + k'e(1-b_i)}{ \sqrt{(d_i+1)(d_j+1)}}\right) \\
&= \frac{k }{ d_i+1} + \frac{\{(1-e)b_ik+e(1-b_i)k'\}d'_i}{ d_i+1} \\
&= \frac{\{(1-e)b_ik+e(1-b_i)k'\}d'_i+k }{ d_i+1}.
\end{flalign}

$\\$

\subsection{Appendix B (entropy between signed and blocked MP)}

We aim to prove that signed MP introduces higher entropy than the blocked ones. Let us assume an ego $i$ with label $k$ and its neighbor node $j$ is connected to $i$ with a signed edge. Firstly, the true label probability ($k$) of node $b$ $\widehat{y}_{b,k}$ increases, while other probabilities $\widehat{y}_{b,o}$ ($o \neq k$) decrease as follows:
\begin{equation}
\begin{gathered}
\widehat{y}^{(t+1)}_p \in
    \begin{cases} \widehat{y}^t_{b,k}-\eta\nabla_b\mathcal{L}_{nll}(Y_i,\widehat{Y}_i)_k > \widehat{y}^t_{b,k} \\
    \widehat{y}^t_{b,o}-\eta\nabla_b\mathcal{L}_{nll}(Y_i,\widehat{Y}_i)_o < \widehat{y}^t_{b,o} \,\,\,\,\, \forall \,\, o\neq k.
    \end{cases}
\end{gathered}
\end{equation}
Now, we analyze the partial derivative $\nabla_b\mathcal{L}_{nll}(Y_i,\widehat{Y}_i)_o$ for $\forall$ $o \neq k$,
\begin{flalign}
&\nabla_b\mathcal{L}_{nll}(Y_i,\widehat{Y}_i)_o 
= \frac{\partial \mathcal{L}_{nll}(Y_i,\widehat{Y}_i)_o }{ \partial \widehat{y}_{b,o}} 
= \frac{\partial \mathcal{L}_{nll}(Y_i,\widehat{Y}_i)_o }{ \partial \widehat{y}_{i,o}} \cdot \frac{\partial \widehat{y}_{i,o} }{ \partial h^{(L)}_{b,o}} \\
&= \frac{1 }{ \widehat{y}_{i,o}}\cdot (\widehat{y}_{i,o}(1-\widehat{y}_{i,o}))
= 1-\widehat{y}_{i,o} > 0,
\end{flalign} 
On the contrary, the gradient of node $s$ has a different sign with node $p$, where we can infer that:
\begin{equation}
\begin{gathered}
\widehat{y}^{(t+1)}_s \in
    \begin{cases} \widehat{y}^t_{s,k}-\eta\nabla_s\mathcal{L}_{nll}(Y_i,\widehat{Y}_i)_k < \widehat{y}^t_{s,k},  \\
    \widehat{y}^t_{s,o}-\eta\nabla_s\mathcal{L}_{nll}(Y_i,\widehat{Y}_i)_o > \widehat{y}^t_{s,o}, \,\, \forall \,\, o\neq k
    \end{cases}
\end{gathered}
\end{equation}
Based on the above analysis, as the training epoch $t$ proceeds, the expectation of signed MP prediction ($\widehat{y}_s$) exhibits higher entropy $H(\mathbb{E}[\widehat{y}_{s}])$ compared to that ($\widehat{y}_b$) of blocked MP $H(\mathbb{E}[\widehat{y}_{b}])$ as below:
\begin{equation}
\label{last_eq}
\begin{gathered}
H(\mathbb{E}[\widehat{y}^{(t+1)}_{s}])-H(\mathbb{E}[\widehat{y}^{(t+1)}_{b}]) > H(\mathbb{E}[\widehat{y}^t_{s}])-H(\mathbb{E}[\widehat{y}^t_{b}]).
\end{gathered}
\end{equation}
$\\$

\subsection{Appendix C (Homophily Estimation)}

As defined in Eq. \ref{homo_est}, our homophily estimation combines the mechanisms of MLP \cite{wang2022powerfulb} and EvenNet \cite{lei2022evennet} as follows:

\begin{equation}
\begin{gathered}
b_i = B_iB_i^T, ,, B := \sigma\left(\sum^L_{l=0} XW^l + \sum^{\lfloor L/2 \rfloor}_{l=0}\tilde{A}^{2l}XW^l\right)
\end{gathered}
\end{equation}

In this equation, the left term $\sum^L_{l=0}XW^l$ employs blocked MP (MLP), while the right term only receives messages from even-hop neighbors (EvenNet). This approach reduces the prediction variance caused by homophily changes while maintaining average performance.

To begin, let us define the $k$-step homophily ratio $\mathbb{H}_k$, following \cite{lei2022evennet}, under \textit{i.i.d.} graphs as follows:

\begin{equation}
\mathbb{H}_k = \frac{1}{N} \sum^{K-1}_{l=0}\left(\Pi^k_{ll} - \sum_{m \neq l}\Pi^k_{lm}\right), ,,, \text{where} ,, \Pi^k = Y^T\tilde{A}^kY
\end{equation}

Since the first term is shared by all $k$-step propagations, the above equation can be rewritten as:

\begin{equation}
\mathbb{H}_k \approx \sum^k_{i=0}\theta_i \mathbb{H}_i(\Pi) = \theta_0 + \theta_1(I - \tilde{L}) + \dots + \theta_k(I - \tilde{L})^k
\end{equation}

Combining the notions of MLP and EvenNet, the equation becomes:

\begin{equation}
\mathbb{H}_0 + \mathbb{H}_{2k} \approx \theta_0 + \left(\theta_0 + \theta_2(I - \tilde{L})^2 + \dots + \theta_{2k}(I - \tilde{L})^{2k} \right)
\end{equation}

By dividing the above equation by 2, we can infer that $\mathbb{E}[\mathbb{H}_0 + \mathbb{H}_{2k}] = \mathbb{E}[\mathbb{H}_{2k}]$, but $Var[\mathbb{H}_0 + \mathbb{H}_{2k}] < Var[\mathbb{H}_{2k}]$. This can enhance heterophily robustness (reduced variance) while maintaining overall performance (expectation).
$\\$

\subsection{Appendix D (Details of Empirical Analysis)}

Figure 1 presents empirical analyses to verify Theorems 10 (left) and 11 (right). For homophily estimation, we use early stopping, saving the predictions with the best validation score if there is no improvement for more than 100 epochs. The edge error estimation follows a slightly different approach, where the \textit{(true)} values are based on the maximum predicted node class as shown below:
\begin{equation}
        \forall (i,j) \in \mathcal{E}, \,\,\, \tilde{\mathcal{E}}^t_{ij} \in
    \begin{cases}
        1, & \,\,\, \widehat{Y}^t_i = \widehat{Y}^t_j\\
        0, & \,\,\, \widehat{Y}^t_i \neq \widehat{Y}^t_j
    \end{cases}
\end{equation}
where $\widehat{Y}^t$ stands for the predicted class at step $t$. The score $s$ can be derived as follows:
\begin{equation}
s = 1\frac{1^T{\sum^N_{i=1}\sum^N_{j=1} \tilde{\mathcal{E}}^t_{ij}}} {|\mathcal{E}|}
\end{equation}
Conversely, the \textit{(pred)} values are derived through Eq. \ref{edge_err_est}. Additional experimental results are provided in Appendix F.
$\\$

\begin{figure*}[t]
 \includegraphics[width=\textwidth]{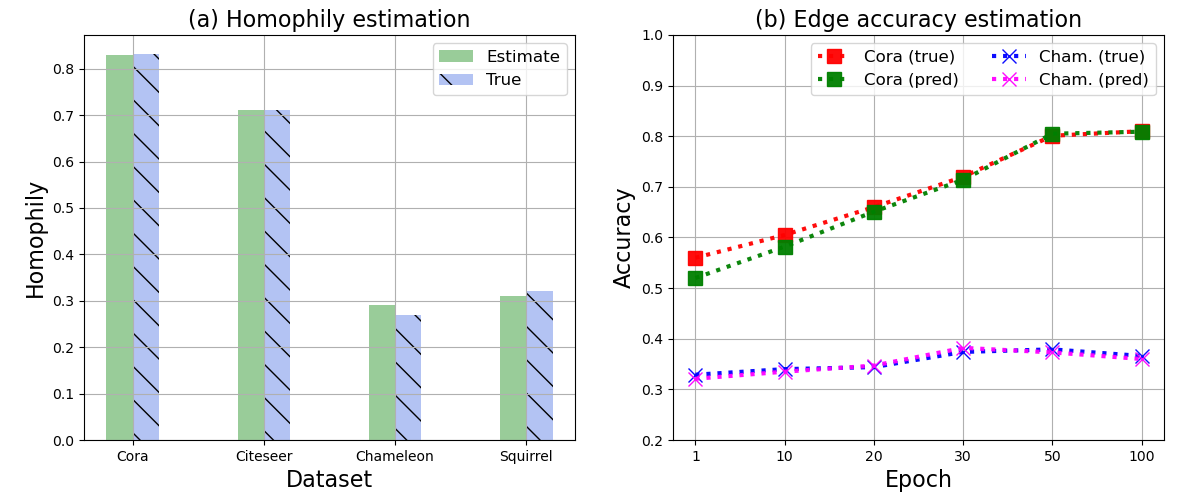}
    \caption{Homophily and edge accuracy (1-error) estimation under \textit{i.i.d.} graphs}
  \label{ablation_emp}
\end{figure*}

\subsection{Appendix E (proof of Theorem 11)}
In Theorem \ref{thm_edg}, we stated that edge error ratio $e_t$ can be inferred using the node classification accuracy of validation sets ($\alpha_{t-1}$) at iteration $t-1$ and the number of classes ($c$). The proof is as follows:
\begin{flalign}
    e_t &= 1 - p(\text{correct edge classification}) \\
    &= 1 - \{ \underset{y_i=y_j}{\underline{\frac{a_{t-1} }{ c} * \frac{a_{t-1} }{ c} * c^2}} + \underset{y_i \neq y_j}{\underline{\frac{1-a_{t-1} }{ c-1} * \frac{1-a_{t-1} }{ c-1} * (c-1)}}\} \\
    &= 1 - \{a^2_{t-1}+\frac{(1-a_{t-1})^2 }{ c-1}\} 
\end{flalign}
If two nodes share the same label ($y_i=y_j$), the edge classification accuracy should be proportional to the product of their validation scores. Conversely, if they have different labels, the edge is correctly classified only if both nodes are incorrectly predicted to have the same label.
$\\$

\subsection{Appendix F (more experiments about Theorem 10 and 11 under \textit{i.i.d.} graphs)}
In addition to Appendix D, we aim to demonstrate that our theorem can estimate the parameters under an \textit{i.i.d.} graph. Following \cite{kothapalli2024neural}, we construct a graph with $N=1,000$ nodes, $c=2$ classes, and $|\mathcal{E}|=10,000$ edges. For each class, the node features are sampled from a Gaussian distribution. Additionally, we fix the degree of each node as $\mathcal{E} / N = 10$ and set the homophily ratio to match the original data. As shown in Figure \ref{ablation_emp}, the predictions are nearly accurate under this condition, validating our analysis of the two theorems.
$\\$

\subsection{Appendix G (Miscellaneous)}
This section introduces several details: (1) time complexity of our model, (2) datasets, (3) implementation, and (4) baselines.
$\\$

\textbf{G.1. Time Complexity of Calibrated GCN}

To begin with, please note that the edge calibration component of our model should be proportional to baseline algorithms such as GPRGNN, FAGCN, or GGCN. Therefore, we focus on the parameter estimation network's complexity, consisting of plain MLP and even-hop propagation networks. Firstly, it is well known that the cost of the MLP follows \(\mathcal{O}(nz(X)F' + F'C)\), where \(nz(\cdot)\) represents the non-zero elements of the inputs, \(F'\) denotes the hidden dimension, and \(C\) is the number of classes. The second part involves even-hop message passing (MP), which can be defined as \(\mathcal{O}(L|\mathcal{E}|\theta_{GCN}/2)\). In summary, the overall cost should be \(\mathcal{O}(nz(X)F' + F'C + L|\mathcal{E}|\theta_{GCN}/2)\), which may require double the computational cost compared to the baseline methods.
$\\$

\textbf{G.2. Datasets}

The statistical details of the datasets are in Table \ref{dataset}.
(1) \textit{Cora, Citeseer, Pubmed} \cite{kipf2016semi} are citation graphs where a node corresponds to a paper, and edges are citations between them. The labels represent the research topics of the papers.    
(2) \textit{Actor} \cite{tang2009social} is a co-occurrence graph of actors appearing in the same movie. The labels represent five types of actors. 
(3) \textit{Chameleon, Squirrel} \cite{rozemberczki2019gemsec} are Wikipedia hyperlink networks. Each node is a webpage, and the edges are hyperlinks. Nodes are categorized into five classes based on monthly traffic.
$\\$

\textbf{G.3. Implementation Details}

All methods, including baselines and ours, are implemented using \textit{PyTorch Geometric}\footnote{\label{code1}https://pytorch-geometric.readthedocs.io/en/latest/modules/nn.html}.
For a fair comparison, we set the hidden dimension for all methodologies to 64. ReLU with dropout is used for non-linearity and to prevent overfitting. We use log-Softmax as the cross-entropy function. The learning rate is set to $1e^{-3}$, and the Adam optimizer is used with a weight decay of $5e^{-4}$. For training, 20 nodes per class are randomly chosen, and the remaining nodes are divided into two parts for validation and testing, following the settings in \cite{kipf2016semi}.
$\\$

\textbf{G.4. Baselines}

\begin{itemize}
    \item \textbf{GCN} \cite{kipf2016semi} is a first-order approximation of Chebyshev polynomials \cite{defferrard2016convolutional}. For all datasets, we simply take 2 layers of GCN.
    \item \textbf{APPNP} \cite{gasteiger2018predict} combines personalized PageRank on GCN. We stack 10 layers and set the teleport probability ($\alpha$) as $\{0.1,0.1,0.1,0.5,0.2,0.3\}$ for Cora, Citeseer, Pubmed, Actor, Chameleon, and Squirrel.
    \item \textbf{GAT} \cite{velickovic2017graph} calculates feature-based attention for edge coefficients. Similar to GCN, we construct 2 layers of GAT. The pair of (hidden dimension, head) is set as (8, 8) for the first layer, while the second layer is (1, \# of classes).
    \item \textbf{GCNII} \cite{chen2020simple} integrates an identity mapping function on APPNP. We set $\alpha=0.5$ and employ nine hidden layers. We increase the weight of identity mapping ($\beta$) that is inversely proportional to the heterophily of the dataset.
    \item \textbf{H$_2$GCN} \cite{zhu2020beyond} suggests the separation of ego and neighbors during aggregation. We refer to the publicly available \textit{source code}\footnote{\label{code2}https://github.com/GemsLab/H2GCN} for implementation. 
    \item \textbf{PTDNet} \cite{luo2021learning} removes disassortative edges before a message-passing. We also utilize the open \textit{source code}\footnote{\label{code3}https://github.com/flyingdoog/PTDNet} and apply confidence calibration.
    \item \textbf{P-reg} \cite{yang2021rethinking} ensembles a regularization term to provide additional information that training nodes might not capture.(\textit{source code}\footnote{https://github.com/yang-han/P-reg})
    \item \textbf{HOG-GCN} \cite{wang2022powerfulb} adaptively controls the propagation mechanism by measuring the homophily degrees between two nodes. (\textit{source code}\footnote{https://github.com/hedongxiao-tju/HOG-GCN})
    \item \textbf{JacobiConv} \cite{wang2022powerfula} studies the expressive power of spectral GNN and establishes a connection with the graph isomorphism testing. (\textit{source code}\footnote{https://github.com/GraphPKU/JacobiConv})
    \item \textbf{GloGNN} \cite{li2022finding} receives information from global nodes, which can accelerate neighborhood aggregation. (\textit{source code}\footnote{https://github.com/RecklessRonan/GloGNN})
    \item \textbf{ACM-GCN} \cite{luan2022revisiting} suggests a local diversification operation through the adaptive channel mixing algorithm. (\textit{source code}\footnote{https://github.com/SitaoLuan/ACM-GNN}) 
    \item \textbf{AERO-GCN} \cite{lee2023towards} improves the deep graph attention to reduce the smoothing effect and improve the performance at deep layers (\textit{source code}\footnote{https://github.com/syleeheal/AERO-GNN}).
    \item \textbf{Auto-HeG} \cite{zheng2023auto} automatically build heterophilic GNN models with search space design, supernet training, and architecture selection \textit{(source code)}\footnote{https://github.com/Amanda-Zheng/Auto-HeG}. 
    \item \textbf{TED-GCN} \cite{yan2024trainable} redefines GCN’s depth $L$ as a trainable parameter, which can control its signal processing capability to model both homophily/heterophily graphs.
    \item \textbf{PCNet} \cite{li2024pc} proposes a two-fold filtering mechanism to extract homophily in heterophilic graphs \textit{(source code)}\footnote{https://github.com/uestclbh/PC-Conv}.
    \item \textbf{GPRGNN} \cite{chien2020adaptive} generalized the personalized PageRank to deal with heterophily and over-smoothing. Referring to the open source \textit{code}\footnote{\label{code4}https://github.com/jianhao2016/GPRGNN}, we tune the hyper-parameters based on the best validation score for each dataset.
    \item \textbf{FAGCN} \cite{bo2021beyond} determines the sign of edges using the node features. We implement the algorithm based on the \textit{sources}\footnote{\label{code5}https://github.com/bdy9527/FAGCN} and tune the hyper-parameters concerning their accuracy.
    \item \textbf{GGCN} \cite{yan2021two} proposes the scaling of degrees and the separation of positive/negative adjacency matrices. We simply take the publicly available \textit{code}\footnote{\label{code6}https://github.com/Yujun-Yan/Heterophily\text{\_}and\text{\_}oversmoothing} for evaluation.
\end{itemize}

\end{appendices}

\end{document}